%% file: example.tex
\documentclass{article}

\usepackage[final]{corl_2026} 

\title{SAGE-3D: Semantic-Aware 3D Representations for Generalizable Vision-Language-Action Models}
\usepackage[utf8]{inputenc} 
\usepackage[T1]{fontenc}    
\usepackage{fontawesome5}
\usepackage{hyperref}       
\usepackage{url}            
\usepackage{booktabs}       
\usepackage{amsfonts}       
\usepackage{nicefrac}       
\usepackage{microtype}      
\usepackage{xcolor}         
\usepackage{graphicx}
\usepackage{tcolorbox}
\usepackage{tabularx}
\newcolumntype{Y}{>{\centering\arraybackslash}X}
\usepackage{makecell}
\usepackage{amsmath}
\usepackage{colortbl,xcolor} 
\usepackage{multirow}
\usepackage{caption}
\usepackage{booktabs}
\usepackage{graphicx}
\usepackage[table]{xcolor}
\usepackage{pifont}
\usepackage{amssymb}
\usepackage{wrapfig}
\usepackage{float}
\newtcolorbox{findingbox}{
    colback=blue!2!white,  
    colframe=black!80,     
    boxrule=0.5pt,         
    arc=3pt,               
    left=4pt,              
    right=4pt,             
    top=4pt,               
    bottom=4pt,            
    boxsep=0pt,            
    width=\linewidth       
}

\definecolor{liberoblue}{RGB}{132,180,214}
\definecolor{calvinyellow}{RGB}{246,226,161}
\newcommand{\methodname}{SAGE-3D}
\author{%
\shortstack[c]{%
\textbf{Pengxiang Ding\textsuperscript{1,2*}, Xuanxuan An\textsuperscript{3*},
Kexian Yu\textsuperscript{4}, Haoying Wang\textsuperscript{3},
Minghui Lin\textsuperscript{2},}\\
\textbf{Xuhao Wan\textsuperscript{4}, Wenxuan Song\textsuperscript{5},
Han Zhao\textsuperscript{1,2}, Yihao Wang\textsuperscript{6},
Donglin Wang\textsuperscript{2,7\textdagger}, Ning Ding\textsuperscript{3\textdagger}}\\[2pt]
\mdseries
\textsuperscript{1}Zhejiang University\quad
\textsuperscript{2}Westlake University\\[-1pt]
\parbox{0.92\linewidth}{%
\centering
\mdseries
\textsuperscript{3}State Key Laboratory of Human-Machine Hybrid Augmented Intelligence,
Institute of Artificial Intelligence and Robotics,
Xi'an Jiaotong University%
}\\[-1pt]
\mdseries
\textsuperscript{4}Zhengzhou University \quad
\textsuperscript{5}HKUST(GZ) \quad
\textsuperscript{6}BAAI \quad
\textsuperscript{7}Westlake Robotics\\[-1pt]
\mdseries
\textsuperscript{*}Equal Contribution \quad
\textsuperscript{\textdagger}Corresponding Authors%
}%
}

\begin{document}
\maketitle
\vspace{-20pt}
\begin{center}
\faGlobe\ \href{https://sage3d-vla.github.io}{sage3d-vla.github.io}
\quad
\faGithub\ \href{https://github.com/OpenHelix-Team/SAGE-3D}{OpenHelix-Team/SAGE-3D}
\end{center}

\begin{abstract}
Vision--Language--Action (VLA) models have shown strong potential for language-conditioned robotic manipulation, but their reliance on 2D visual backbones limits precise reasoning about depth, spatial relations, and contact-rich actions. We propose \methodname{}, a semantic-aware 3D VLA framework that aligns geometric structure with task semantics for instruction-conditioned action prediction. Motivated by representation analysis, our key insight is that effective 3D VLA policies require both transferable geometric cues and instruction-relevant semantic grounding: geometry should not be injected as an isolated modality, but organized around objects, relations, and affordances relevant to the task. To realize this, \methodname{} introduces a \textbf{3D Semantic-Aware Feature Generator} (SAFG), which integrates RGB semantics, depth information, and task instructions to produce \textbf{3D Reasoning Tokens}. These tokens are injected into the action head through a FiLM-based geometry-conditioned decoder, allowing semantic-aware spatial context to directly modulate action generation. Extensive experiments in simulation and real-world manipulation show that \methodname{} improves overall benchmark performance and training efficiency, with the largest gains on spatially demanding tasks. Notably, \methodname{} reaches \textbf{100\%} success on \textsc{LIBERO-Spatial}, establishing a new state-of-the-art result and, to our knowledge, the first result to saturate this benchmark.
\end{abstract}

\begin{figure}[htbp]
  \centering
  \includegraphics[width=1.0\linewidth]{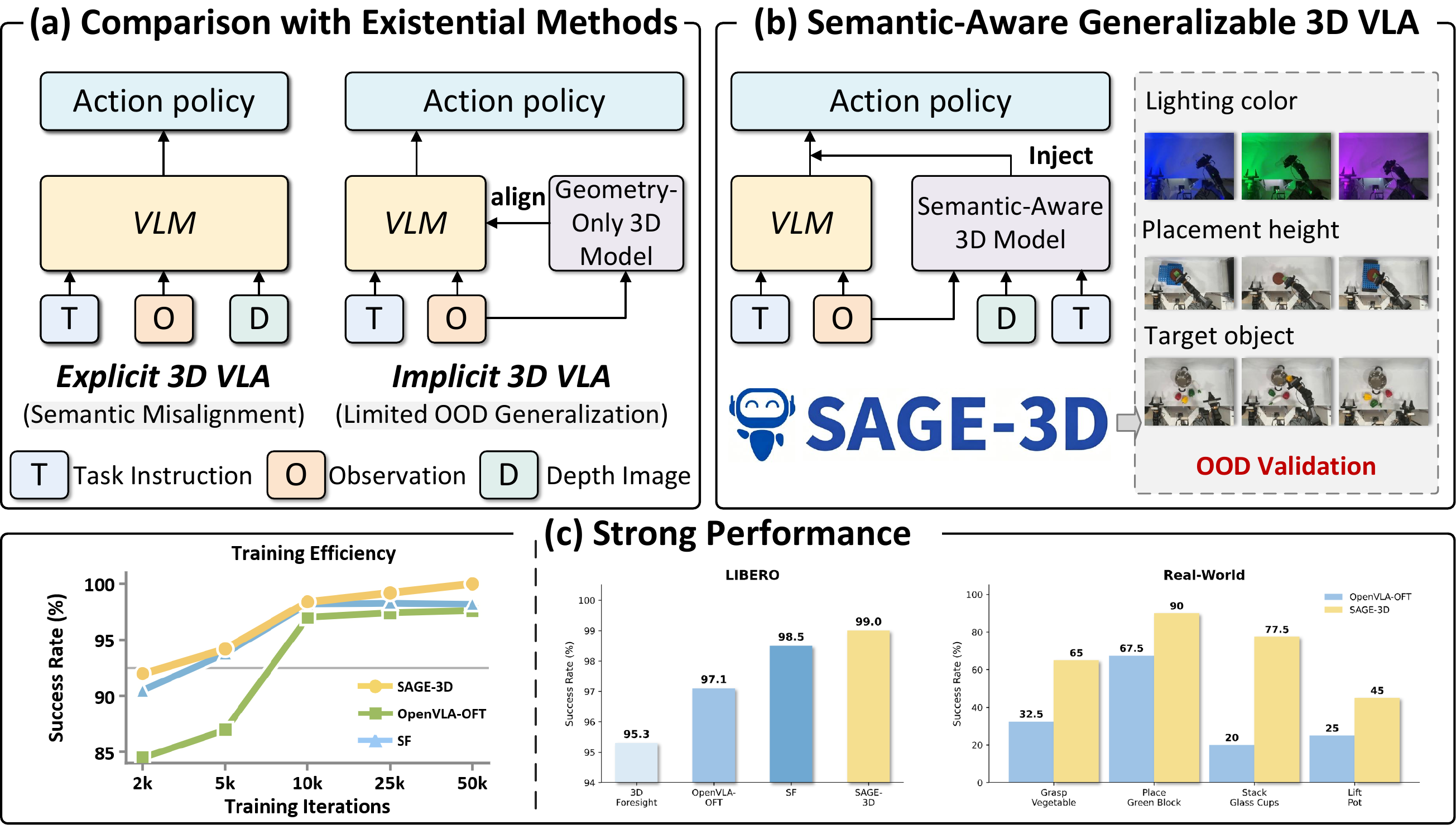}
  \caption{\textbf{Overview of the proposed SAGE-3D.} Compared to (a) existing methods, (b) SAGE-3D integrates a semantic-aware 3D foundation model to couple 3D semantics and spatial geometry with strong OOD generalization. (c) Results in both simulation and the real world show superior success.} 
  \label{fig:teaser}
  \vspace{-15pt}
\end{figure}
\input{sec/1_intrc}
\input{sec/2_rev}
\input{sec/3_method}
\input{sec/4_exper}

\input{sec/5_relat}
\input{sec/7_con}


\clearpage
\acknowledgments{This work is supported by the Brain Science and Brain-like Intelligence Technology — National Science and Technology Major Project (Grant No. 2022ZD0208800), National Natural Science Foundation of China (Grant No. 62495092, 62495090).
}


\bibliography{example}  
\input{sec/appendix}

\end{document}

%% file: sec/1_intrc.tex
\section{Introduction}

Vision--Language--Action (VLA) models have emerged as a promising paradigm for language-conditioned robotic manipulation, mapping visual observations and natural-language instructions to robot actions~\cite{VLA,OpenVLA,openvla-oft,bu2025univla,pi0}. Yet many manipulation tasks require reliable 3D geometric grounding for depth, spatial relations, contact, and precise end-effector motion. Since most existing VLAs rely on 2D visual backbones, their limited 3D understanding constrains precision and robustness in geometry-sensitive tasks.

Recent work incorporates 3D information into VLA policies through two representative strategies~\cite{bhat20253d,3DForesight,4D-VLA,DepthVLA,fan2026any3d}. \emph{Explicit 3D fusion} augments visual representations with depth, point clouds, or geometry-aware features~\cite{4D-VLA,fan2026any3d,qu2025spatialvla,sun2025geovla}. These features provide useful metric cues, but they are not necessarily aligned with the language-conditioned semantic space used by the policy; a model may observe accurate geometry while failing to identify the instruction-relevant objects or relations. \emph{Implicit 3D injection} instead supervises intermediate VLA representations with geometry-aware teacher features~\cite{zhang2025inspire,spatialforcing}. This can improve data efficiency by embedding 3D cues into semantic visual tokens, but the learned geometry can remain tied to the teacher and training distribution, weakening transfer under visual noise, domain shift, and real-world deployment. These limitations indicate that 3D-aware VLA policies require both transferable geometry and task-relevant semantic alignment.

We revisit these paradigms through representation analysis. Depth probing shows that implicit alignment struggles to preserve decodable depth under noise, out-of-distribution scenes, and real-world observations, whereas explicit geometric representations provide more stable depth structure. However, attention analysis shows that geometry-centric tokens can attend to geometrically salient but task-irrelevant regions, indicating limited instruction-object coupling. These findings motivate \emph{semantic-aware geometric alignment}: 3D structure should not be injected as an isolated modality, but organized around task-relevant semantics and spatial relations.

To this end, we propose \methodname{}, an explicit semantic-aware 3D VLA framework. Its core component is a \textbf{3D Semantic-Aware Feature Generator} (SAFG), which integrates RGB semantics, depth information, and task instructions to produce \textbf{3D Reasoning Tokens}. These tokens encode relative object distances, scene layout, and affordance-related spatial relations for instruction-conditioned action prediction. Instead of concatenating 3D features with visual tokens, \methodname{} injects the reasoning tokens into the action head through a FiLM-based geometry-conditioned decoder, enabling semantic-aware spatial context to modulate action generation directly. Our main contributions are summarized as follows:
\vspace{-1pt}
\begin{itemize}
    \item We systematically compare explicit 3D fusion and implicit 3D injection, showing that effective 3D-VLA policies require both transferable geometry and task-relevant semantics.
    \vspace{-1pt}
    \item We propose \textbf{SAGE-3D}, a 3D VLA framework whose Semantic-Aware Feature Generator (SAFG) produces instruction-conditioned reasoning tokens for action prediction.
\vspace{-1pt}
    \item Extensive simulated and real-world experiments demonstrate improved performance and training efficiency. SAGE-3D achieves \textbf{100\%} success on \textsc{LIBERO-Spatial}~\cite{libero}, establishing a new state-of-the-art result.
\end{itemize}

%% file: sec/2_rev.tex
\section{Revisiting 3D VLA paradigms through representation analysis}
\vspace{-5pt}
Although recent 2D VLA models have advanced robotic manipulation, robust 3D geometric understanding and instruction-relevant spatial reasoning remain challenging. This section examines how different 3D representations are preserved and used in VLA models, addressing two questions.

\textbf{\textit{Question 1. Which type of 3D prior better preserves transferable geometry: implicit alignment or explicit representation?}}


\textbf{Setup.}
We probe whether intermediate VLA features retain decodable geometry across domains. For the implicit paradigm, we use frozen SF~\cite{spatialforcing} representations fine-tuned on \textsc{LIBERO-Spatial}~\cite{libero}, and train a lightweight DPT~\cite{DPT} prediction head on the SF-aligned 24th VLA layer and its three preceding layers, with a two-layer MLP for feature alignment. For the explicit paradigm, we adopt GeoVLA~\cite{sun2025geovla} and use VGGT~\cite{vggt} as a geometry foundation model reference. We evaluate all methods on four scenarios: LIBERO in-domain (ID), noisy LIBERO scenes, out-of-distribution (OOD) scenes from CALVIN, and real-world scenes, and use ZoeDepth~\cite{zoedepth} predictions as pseudo-GT depth.

\begin{figure}[H]
\vspace{-10pt}
  \centering
    \includegraphics[width=1.0\linewidth]{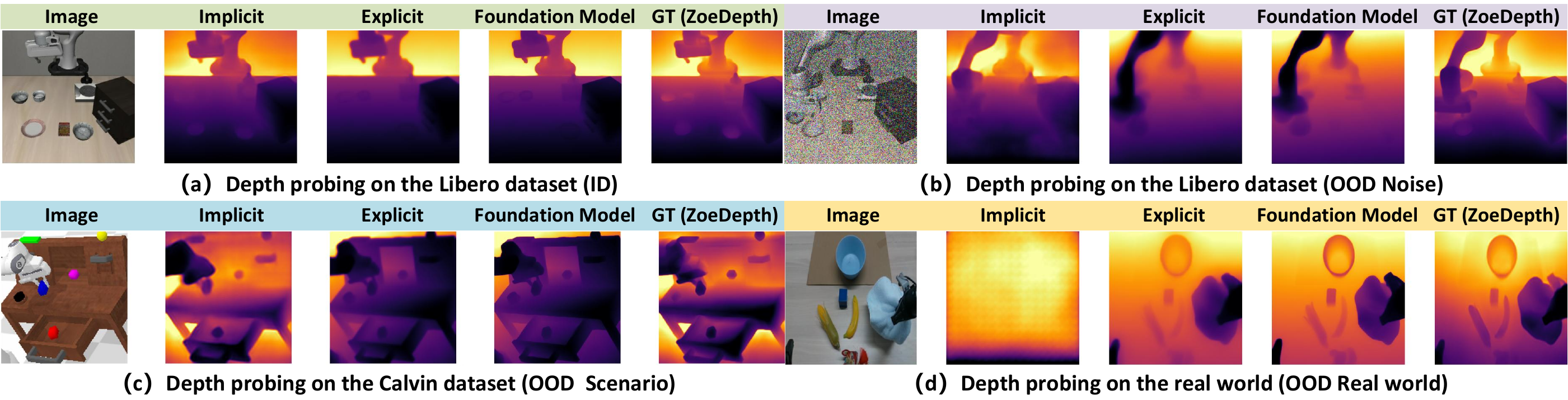}
  \caption{\textbf{Depth probing across validation scenarios.} 
Comparison of implicit SF, explicit GeoVLA, and VGGT depth predictions against pseudo-GT across ID, noisy, OOD, and real-world scenarios.
  }
  \label{fig:example2}
\vspace{-15pt}
\end{figure}


\textbf{Analysis.} Fig.~\ref{fig:example2} shows that frozen SF embeddings degrade under noise, OOD, and real-world shifts. In contrast, explicit-geometry methods, including GeoVLA and the VGGT 3D foundation model, preserve more stable depth, supporting explicit representations as transferable geometric priors.

\vspace{-0.2em}
\begin{findingbox}
\textbf{Finding 1.} \textit{Compared with SF-style implicit alignment, explicit geometric representations preserve more transferable depth structure, suggesting their value as robust geometric anchors.}
\end{findingbox}

However, geometric anchoring alone does not guarantee effective manipulation, because the policy must still identify instruction-relevant objects and relations.

\vspace{-0.1em}
\textbf{\textit{Question 2. What remains missing from explicit geometry-centric 3D representations?}}

\textbf{Setup.}
Although many 3D tokens encode metric geometry, they may overlook object affordances and instruction-relevant relations. We compare three frozen feature extractors on \textsc{LIBERO-Spatial} under the same backbone, feature-injection strategy, and training setup: (i) Uni3D~\cite{uni3d}, whose point-cloud features are projected onto the image plane for fair comparison; (ii) VGGT, a geometry-oriented baseline using patch features extracted from its 24th layer; and (iii) SAFG, introduced in Section~\ref{model_arch}, which integrates visual semantics with metric depth to model object distances and scene layout.

\begin{figure}[H]
\vspace{-10pt}
  \centering
  \includegraphics[width=1.0\linewidth]{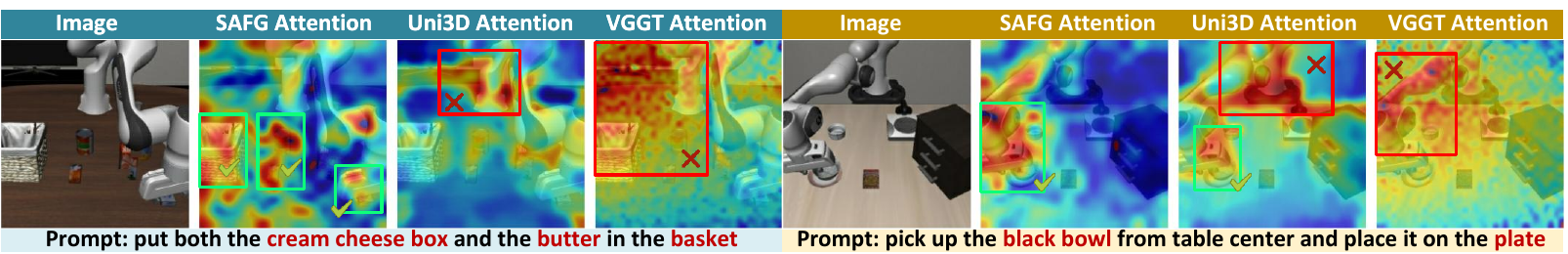}
  
  \caption{\textbf{Attention visualization of 3D feature representations.} We visualize the action-to-observation attention maps produced by VLA policies trained with different 3D feature representations under the same backbone and training setup.
  }
  \label{fig:example3}
  \vspace{-15pt}
\end{figure}

\textbf{Analysis.}
As shown in Fig.~\ref{fig:example3}, the policy trained with SAFG features focuses more on instruction-relevant regions, whereas those trained with Uni3D and VGGT features attend more broadly to visually or geometrically salient regions, including task-irrelevant areas. Although Uni3D provides semantically aligned 3D features and VGGT offers robust geometric cues, SAFG better grounds spatial relations in task semantics and affordances, benefiting from more diverse pre-training data.

\begin{findingbox}
\textbf{Finding 2.} \textit{Compared with geometry-centric tokens, semantically aligned 3D representations focus attention more on instruction-relevant objects, suggesting stronger instruction-object coupling.}
\end{findingbox}

%% file: sec/3_method.tex
\section{Method}

\subsection{Preliminary}
VLA models are designed to map multimodal observations into executable actions. Formally, given the observation $\mathcal{O}_t \in \mathbb{R}^{H \times W \times 3}$, the task instruction $\mathcal{T}$, and the proprioceptive state $\mathcal{P}_t$ at time step $t$, the agent predicts a future $N$-step action chunk $A_{t:t+N-1} = [a_t, a_{t+1}, \dots, a_{t+N-1}]$. Here, each single-step action $a_i \in \mathbb{R}^7$ represents the 3D end-effector pose change and gripper state. 
To incorporate 3D spatial priors, we decouple the policy input into VLM semantic features and depth-aware spatial reasoning tokens. 
Specifically, the visual encoder and LLM extract perception tokens $Z_P$ from the observation, while the \textbf{3D Semantic-Aware Feature Generator} (SAFG) produces \textbf{3D Reasoning Tokens} $Z_{\mathrm{RT}}$ from visual, depth, and task inputs. These tokens provide geometric cues as auxiliary 3D priors for the policy. The action chunk is then predicted conditioned on $Z_P$, $Z_{\mathrm{RT}}$, $\mathcal{T}$, and the proprioceptive state: \(A_{t:t+N-1} = \pi(Z_P, Z_{\mathrm{RT}}, \mathcal{P}_t)\).

\subsection{Model Architecture}
\label{model_arch}

\begin{figure}[tb]
  \centering
  \includegraphics[width=1.0\linewidth]{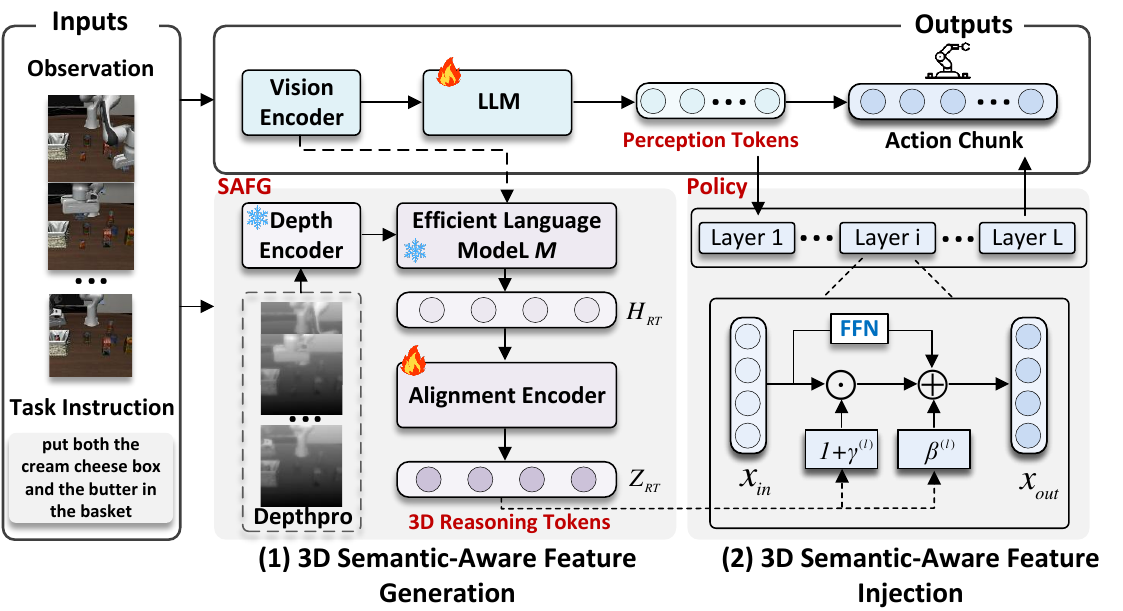}
  \caption{\textbf{Architecture of \methodname{}.} The VLA backbone predicts action chunks from visual observations and task instructions. In parallel, SAFG generates 3D reasoning tokens from visual, depth, and task inputs; these tokens are injected into the action head for action generation.
  }
  \label{fig:example4}
  \vspace{-10pt}
\end{figure}

\textbf{3D Semantic-Aware Feature Generation.} 
As shown in Fig.~\ref{fig:example4}, SAFG generates semantic-aware 3D reasoning representations, following the spatial reasoning design of MIDI~\cite{liu2025ssr}. Given the observation $\mathcal{O}_t$, DepthPro~\cite{bochkovskiy2025depth} estimates the depth map \(\mathcal{D}_t=\Phi_{\mathrm{depth}}(\mathcal{O}_t)\). We then extract RGB semantic features $Z_V=\mathcal{E}_{\mathrm{vis}}(\mathcal{O}_t)$ and geometric depth features $Z_D=\phi_{\mathrm{depth}}(\mathcal{D}_t)$.
Together with the instruction $\mathcal{T}$, these features are fed into an efficient language model Mamba~\cite{mamba} \(\mathcal{M}\), whose hidden states capture spatial relations and affordances. We project the last-layer hidden states:

\vspace{-10pt}
\begin{equation}
H_{\mathrm{RT}} = \phi_{\mathrm{proj}}(\mathrm{LastHidden}(\mathcal{M}(Z_V, Z_D, \mathcal{T}))) \in \mathbb{R}^{L \times D_{\mathrm{RT}}}
\end{equation}
where $H_{\mathrm{RT}}$ denotes the projected hidden representations from the last layer of \(\mathcal{M}\). 
Since these representations lie outside the action decoding space of the VLA, the alignment module maps them into $\mathbb{R}^{D_{\mathrm{act}}}$ through a nonlinear projection:
\vspace{-0.1em}
\begin{equation}
Z_{\mathrm{RT}} = \mathrm{MLP}_{\mathrm{align}}(H_{\mathrm{RT}}) \in \mathbb{R}^{L \times D_{\mathrm{act}}}
\end{equation}
where $Z_{\mathrm{RT}}$ denotes semantic-aware 3D reasoning tokens that provide geometric priors for the policy. $\mathrm{MLP}_{\mathrm{align}}$ consists of two linear layers and a GELU activation function. 

\textbf{3D Semantic-Aware Feature Injection.}
To inject spatial context into action generation, we design a FiLM-based ~\cite{film} geometry-conditioned decoder. Specifically, an aggregated descriptor of $Z_{\mathrm{RT}}$ predicts the modulation parameters $\gamma$ and $\beta$, which scale and shift the action-head input feature $x_{\mathrm{in}}$. The modulated feature is then processed by a feed-forward network with a residual connection:
\begin{equation}
x_{\mathrm{out}} = \mathrm{FFN}((1 + \gamma) \odot x_{\mathrm{in}} + \beta) + x_{\mathrm{in}} .
\end{equation}
For the first layer, $x_{\mathrm{in}}$ is obtained by applying LayerNorm and an MLP to the perception token $Z_P$. After the final decoder layer, $x_{\mathrm{out}}$ is further passed through LayerNorm and an MLP to predict the action chunk $\hat{A}_{t:t+N-1}$.

\textbf{Training Objective.} We train the policy end-to-end with an $L_1$ regression loss to predict continuous action vectors. The objective minimizes the absolute deviation between the predicted action chunk and the expert action chunk $A_{t:t+N-1}$: $\mathcal{L}_{\mathrm{action}} = \mathbb{E}[\| A_{t:t+N-1} - \hat{A}_{t:t+N-1} \|_1]$.

%% file: sec/4_exper.tex
\section{Experiments}
We evaluate \methodname{} by answering the following research questions (RQs):
\textbf{RQ1}: How much does \methodname{} improve performance in in-domain scenarios?
\textbf{RQ2}: How well does \methodname{} generalize to novel environments?
\textbf{RQ3}: Which components of \methodname{} contribute most to its effectiveness?
\textbf{RQ4}: How well does the proposed approach transfer to real-world execution?

\subsection{Evaluation on In-Domain Scenarios (RQ1)}
\label{sec:in-domain}

\noindent\textbf{Experimental Setup.}
We evaluate on the LIBERO benchmark~\cite{libero}, following prior protocols~\cite{openvla-oft, pi0}.
The agent observes RGB images and receives natural-language instructions to control a simulated robotic arm.
LIBERO includes four task suites: \textsc{LIBERO-Spatial}, \textsc{LIBERO-Object}, \textsc{LIBERO-Goal}, and \textsc{LIBERO-Long}.
We consider this setting \emph{in-domain} because the train/test split shares the same background and scene layout, with only minor object perturbations.

\noindent\textbf{Metrics and Baselines.}
We report (i) final success rate (SR) and (ii) training efficiency, measured by SR at the same number of training iterations. We compare against (i) \emph{standard 2D VLA} baselines (e.g., $\pi_0$~\cite{pi0}, OpenVLA-OFT~\cite{openvla-oft}),
(ii) \emph{implicit 3D VLA} methods (SF~\cite{spatialforcing}, InSpire~\cite{zhang2025inspire}),
and (iii) \emph{explicit 3D VLA} methods (e.g., GeoVLA~\cite{sun2025geovla}, 4D-VLA~\cite{4D-VLA}).

\begin{table*}[t]
  \centering
  \scriptsize

  \caption{\textbf{Performance on LIBERO compared with state-of-the-art methods.}
  Best results are in \textbf{bold}, and second-best results are \underline{underlined}.
  Missing results are denoted by ``-''.}

  \label{tab:libero}

  \setlength{\tabcolsep}{4pt}
  \renewcommand{\arraystretch}{1.0}

  \begin{tabularx}{0.98\linewidth}{lYYYYY}
    
    \toprule

    \multirow{2}{*}{Method}
& \multicolumn{5}{c}{\textbf{LIBERO}}
\\

\cmidrule(lr){2-6}

&
\makecell[c]{Spatial\\ SR (\%)}
&
\makecell[c]{Object\\ SR (\%)}
&
\makecell[c]{Goal\\ SR (\%)}
&
\makecell[c]{Long\\ SR (\%)}
&
\makecell[c]{Average\\ SR (\%)}
\\

    \midrule

    \rowcolor{black!4}
    \multicolumn{6}{c}{\textbf{Standard 2D VLA}}
    \\

    \addlinespace[2pt]

    TraceVLA \cite{Tracevla} {\scriptsize\textit{[ICLR'25]}}
    & 84.6 & 85.2 & 75.1 & 54.1 & 74.8
    \\

    OpenVLA \cite{OpenVLA} {\scriptsize\textit{[CoRL'24]}}
    & 84.7 & 88.4 & 79.2 & 53.7 & 76.5
    \\

    OpenVLA-OFT \cite{openvla-oft} {\scriptsize\textit{[RSS'25]}}
    & 97.6 & 98.4 & 97.9 & 94.5 & 97.1
    \\

    UniVLA \cite{bu2025univla} {\scriptsize\textit{[RSS'25]}}
    & 96.5 & 96.8 & 95.6 & 92.0 & 95.2
    \\

    CoT-VLA \cite{GraphCoT-VLA} {\scriptsize\textit{[CVPR'25]}}
    & 87.5 & 91.6 & 87.6 & 69.0 & 83.9
    \\

    $\pi_0$-FAST \cite{pi-fast} {\scriptsize\textit{[RSS'25]}}
    & 96.4 & 96.8 & 88.6 & 60.2 & 85.5
    \\

    $\pi_0$ \cite{pi0} {\scriptsize\textit{[RSS'25]}}
    & 96.8 & 98.8 & 95.8 & 85.2 & 94.2
    \\

    VLA-Adapter \cite{vla-adapter} {\scriptsize\textit{[AAAI'26]}}
    & 97.8 & 99.2 & 97.2 & 95.0 & 97.3
    \\

    \addlinespace[4pt]

    \midrule

    \rowcolor{black!4}
    \multicolumn{6}{c}{\textbf{Implicit 3D VLA}}
    \\

    \addlinespace[2pt]

    InSpire \cite{zhang2025inspire} {\scriptsize\textit{[arXiv'25]}}
    & 90.7 & 94.3 & 88.3 & 73.3 & 86.7
    \\

    SF \cite{spatialforcing} {\scriptsize\textit{[ICLR'26]}}
    & \underline{99.4} & \underline{99.6}
    & \underline{98.8} & 96.0
    & \underline{98.5}
    \\

    \addlinespace[4pt]

    \midrule

    \rowcolor{black!4}
    \multicolumn{6}{c}{\textbf{Explicit 3D VLA}}
    \\

    \addlinespace[2pt]

    SpatialVLA \cite{qu2025spatialvla} {\scriptsize\textit{[RSS'25]}}
    & 88.2 & 89.9 & 78.6 & 55.5 & 78.1
    \\

    DepthVLA \cite{DepthVLA} {\scriptsize\textit{[arXiv'25]}}
    & 96.4 & 98.0 & 95.8 & 89.2 & 94.9
    \\

    4D-VLA \cite{4D-VLA} {\scriptsize\textit{[arXiv'25]}}
    & 88.9 & 95.2 & 90.0 & 79.1 & 88.6
    \\

    Any3D-VLA \cite{fan2026any3d} {\scriptsize\textit{[arXiv'26]}}
    & 75.6 & 77.2 & 80.2 & 40.8 & 68.5
    \\

    GeoVLA \cite{sun2025geovla} {\scriptsize\textit{[arXiv'25]}}
    & 98.4 & 99.0 & 96.6 & \underline{96.6} & 97.7
    \\

    3D-CAVLA \cite{bhat20253d} {\scriptsize\textit{[CVPR'25]}}
    & 98.2 & \textbf{99.8} & 98.2 & 96.1 & 98.1
    \\

    3D Foresight \cite{3DForesight} {\scriptsize\textit{[ICRA'26]}}
    & 96.4 & 98.0 & 94.8 & 92.0 & 95.3
    \\

    \addlinespace[2pt]

   \cellcolor{blue!6}\textbf{\methodname{} (Ours)}
&
\cellcolor{blue!6}\textbf{100.0}
&
\cellcolor{blue!6}\textbf{99.8}
&
\cellcolor{blue!6}\textbf{99.2}
&
\cellcolor{blue!6}\textbf{97.0}
&
\cellcolor{blue!6}\textbf{99.0}
\\

    \bottomrule

  \end{tabularx}

  \vspace{-15pt}

\end{table*}

\textbf{Success Rate Analysis.}
Table~\ref{tab:libero} shows that \methodname\ achieves strong performance across all four LIBERO suites. Specifically, it attains SR of \textbf{100.0}/\textbf{99.8}/\textbf{99.2}/\textbf{97.0} on \textsc{Spatial}/\textsc{Object}/\textsc{Goal}/\textsc{Long}, yielding an average SR of \textbf{99.0}\%. Notably, \methodname\ ranks first on every suite and the overall average, indicating that the gains are not confined to a particular task category but generalize across diverse in-domain settings. The improvement is particularly evident on the long-horizon \textsc{LIBERO-Long} suite, where error accumulation is more pronounced and sustained closed-loop execution is required.

\begin{wrapfigure}{r}{0.42\linewidth}
    \vspace{-10pt}
    \centering
    \includegraphics[width=\linewidth]{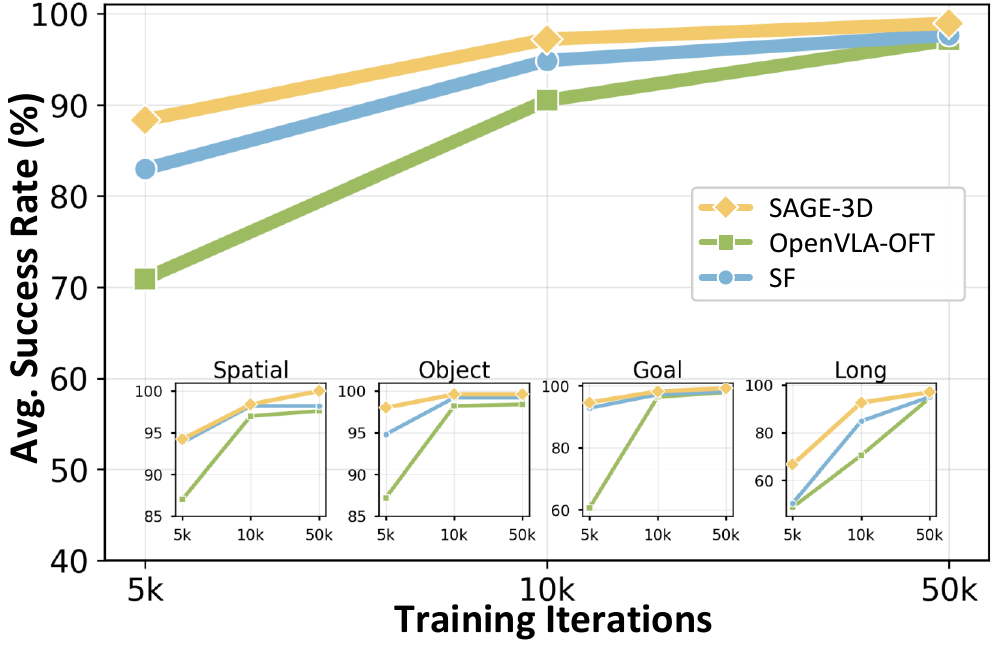}
    \caption{\textbf{Training efficiency across four task suites on LIBERO.}}
    \label{fig:training_efficiency}
    \vspace{-12pt}
\end{wrapfigure}

\noindent\textbf{Training Efficiency Analysis.}
Fig.~\ref{fig:training_efficiency} shows that \methodname\ is more training-efficient on LIBERO.
With the same number of training iterations, \methodname\ consistently achieves higher success rates than OpenVLA-OFT and SF across all suites, indicating improved sample efficiency and faster convergence.
The gains are most pronounced on the more challenging \textsc{Goal} and long-horizon \textsc{Long} tasks, where \methodname\ attains strong performance early and converges to a higher plateau as training scales from 10k to 50k iterations.
\textbf{We attribute this behavior to our joint semantic-and-depth representations, which reduce redundant features and facilitate more efficient optimization.}

\subsection{Evaluation on Out-of-Domain Scenarios (RQ2)}
\label{sec:out-of-domain}
\vspace{-5pt}

\noindent
\begin{minipage}[t]{0.48\linewidth}
\vspace{0pt}

\textbf{Experimental Setup.}
We evaluate on the LIBERO-Pro benchmark~\cite{libero-pro}, which extends LIBERO with task-level distribution shifts. 
For each task suite, we choose four perturbation dimensions: Object Attribute (Obj), Initial Position (Pos), Semantic (Sem), and Task (Task). 
Following the zero-shot protocol, models are trained on LIBERO and evaluated directly on unseen LIBERO-Pro tasks without additional fine-tuning, providing a testbed for generalization under task and spatial variations.

\end{minipage}
\hfill
\begin{minipage}[t]{0.48\linewidth}

\centering
\scriptsize

\captionof{table}{Zero-shot performance on LIBERO-Pro under four perturbation dimensions.
Best results are highlighted in \textbf{bold}.}

\label{tab:libero_pro}

\setlength{\tabcolsep}{3pt}
\renewcommand{\arraystretch}{1.0}
\begin{tabularx}{\linewidth}{lYYYYY}
\toprule
Method
& Spatial
& Object
& Goal
& Long
& Average$\uparrow$
\\
\midrule

OpenVLA-OFT
&0.27&0.41&0.14&0.11&0.23
\\

MolmoAct
&0.45&\textbf{0.49}&0.38&0.34&0.42
\\

\midrule

\rowcolor{blue!8}
\textbf{SAGE-3D (Ours)}
&\textbf{0.49}
&0.48
&\textbf{0.45}
&\textbf{0.35}
&\textbf{0.44}
\\

\rowcolor{gray!10}
$\triangle$ vs. OpenVLA-OFT
&+0.22
&+0.07
&+0.31
&+0.24
&+0.21
\\

\bottomrule
\end{tabularx}

\end{minipage}


\noindent\textbf{Metrics and Baselines.}
We report success rates across four perturbation dimensions and the average success rate (0.00–1.00).

\textbf{Success Rate Analysis.}
Table~\ref{tab:libero_pro} reports zero-shot results on LIBERO-Pro. Relative to the baseline OpenVLA-OFT, SAGE-3D raises the average success rate from 0.23 to 0.44, achieving a notable absolute improvement of 0.21. 
More specifically, SAGE-3D surpasses OpenVLA-OFT by 0.22 on Spatial, 0.07 on Object, 0.31 on Goal, and 0.24 on Long. 
The largest gains on Goal and Long indicate that our semantic-aware 3D representation is particularly beneficial under more challenging distribution shifts, where robust task understanding and spatial reasoning are both required.

\vspace{-5pt}
\subsection{Component-wise Analysis (RQ3)}
\label{sec:component}
\vspace{-5pt}

\textbf{Experimental Setup.} We conduct ablation studies on the challenging long-horizon \textsc{LIBERO-Long} suite. All variants use the same training setup, VLM backbone, differing only in how spatial features are integrated into the VLM.

\begin{figure}[tb]
  \centering
  \includegraphics[width=1.0\linewidth]{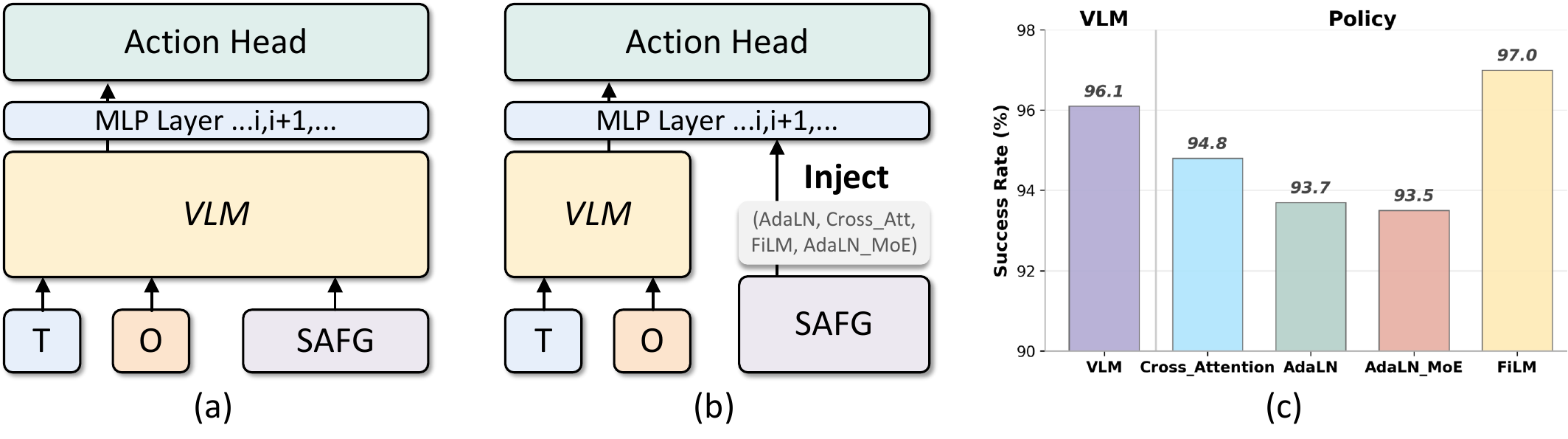}
  \caption{\textbf{Spatial information injection strategies.} We compare (a) VLM-side and (b) policy-side $Z_\mathrm{RT}$ injection, and evaluate different fusion mechanisms in (c).
  }
  \label{fig:injection_policy}
  \vspace{-15pt}
\end{figure}

\textbf{Spatial information injection strategies.} We evaluate two mainstream explicit depth-injection pathways: (i) injecting depth features into the VLM branch (e.g., Evo-0~\cite{lin2025evo}), and (ii) injecting them into the policy (e.g., LAST0~\cite{last0}, ChatVLA~\cite{zhou2025chatvla}). As shown in Fig.~\ref{fig:injection_policy} (c), injecting $Z_\mathrm{RT}$ into the VLM branch achieves 96.1\% success, while policy-side injection with FiLM performs best, reaching 97.0\%. These results suggest that SAFG features are more effective when used to modulate the policy branch, and that \textbf{FiLM} provides the most suitable fusion mechanism.

\begin{figure}[tb]
  \centering
  \includegraphics[width=1.0\linewidth]{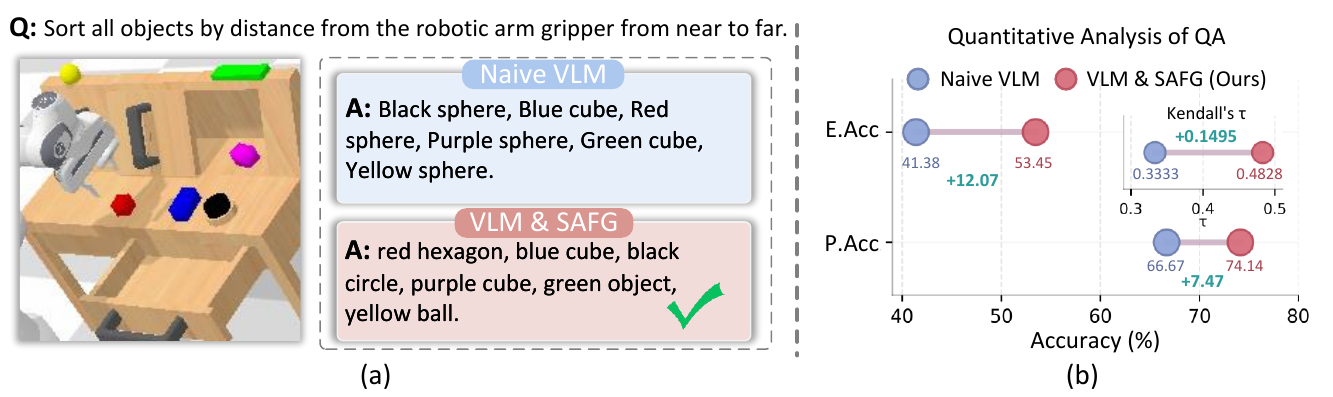}
  \caption{\textbf{Zero-shot spatial understanding with SAFG.} VLM\&SAFG improves gripper-distance sorting over a naive VLM, with an example in (a) and quantitative results in (b).}
  \label{fig:QA}
  \vspace{-10pt}
\end{figure}

\textbf{Zero-shot 3D Spatial Reasoning.}
We evaluate the frozen SAFG on zero-shot spatial QA without fine-tuning or demonstrations.
Using 500 uniformly sampled frames from CALVIN ABC$\rightarrow$D, we retain objects meeting a fixed visibility criterion and derive ground-truth distance orderings from simulator-provided 3D centers.  The task ranks tabletop objects by gripper distance. As shown in Fig.~\ref{fig:QA}(a), VLM\&SAFG produces more consistent rankings than naive VLM, improving accuracy, pairwise order accuracy, and Kendall's $\tau$~\cite{kendall} by $+12.07\%$, $+7.47\%$, and $+0.1495$, respectively. This suggests that SAFG-generated $Z_\mathrm{RT}$ provides 3D priors by integrating depth with visual semantics.

\begin{wraptable}[7]{r}{0.40\linewidth}
\vspace{-12pt}
\centering
\small
\caption{\textbf{Ablation on the role of SAFG.}  Frozen SAFG keeps the alignment encoder trainable. Best results are in \textbf{bold}.}
\label{tab:midi}
\vspace{1mm}
\begin{tabular}{lcc}
\toprule
Setting & $Z_\mathrm{RT}$ & SR (\%)$\uparrow$ \\
\midrule
w/o SAFG & \textcolor{red}{\ding{55}} & 94.5 \\
w/ Trainable SAFG & \textcolor{green!60!black}{\textbf{\checkmark}} & 95.4 \\
\rowcolor{blue!6}
w/ Frozen SAFG  & \textcolor{green!60!black}{\textbf{\checkmark}} & \textbf{97.0} \\
\bottomrule
\end{tabular}
\vspace{-10pt}
\end{wraptable}

\noindent\textbf{Ablation on SAFG.}
We analyze the role of SAFG in \methodname{} in Table~\ref{tab:midi}. Removing SAFG yields 94.5\% success, while trainable SAFG improves it to 95.4\%, showing the benefit of SAFG's $Z_\mathrm{RT}$ representations. Freezing SAFG performs best at 97.0\%, suggesting that fixed semantic-geometric priors improve generalization, whereas finetuning may cause representation drift.

\subsection{Real-world Experiments (RQ4)}
\label{sec:real-world}

\textbf{Experimental Setup.} 
We evaluate \methodname\ on an AgileX real-robot setup using four tasks that cover different dimensions of spatial generalization: \textbf{grasp right-side vegetable under target object variations}, \textbf{place green block under height variations}, \textbf{stack glass cups under lighting variations}, and \textbf{a dual-arm pot-lifting task}. All tasks are jointly trained with 50 trajectories per task. Each single-arm task is tested on four variations with 10 trials each, and  20 trials for the dual-arm task. We report the task success rate (SR) as the evaluation metric.

\begin{figure}[tb]
\vspace{-10pt}
  \centering
  \includegraphics[width=1.0\linewidth]{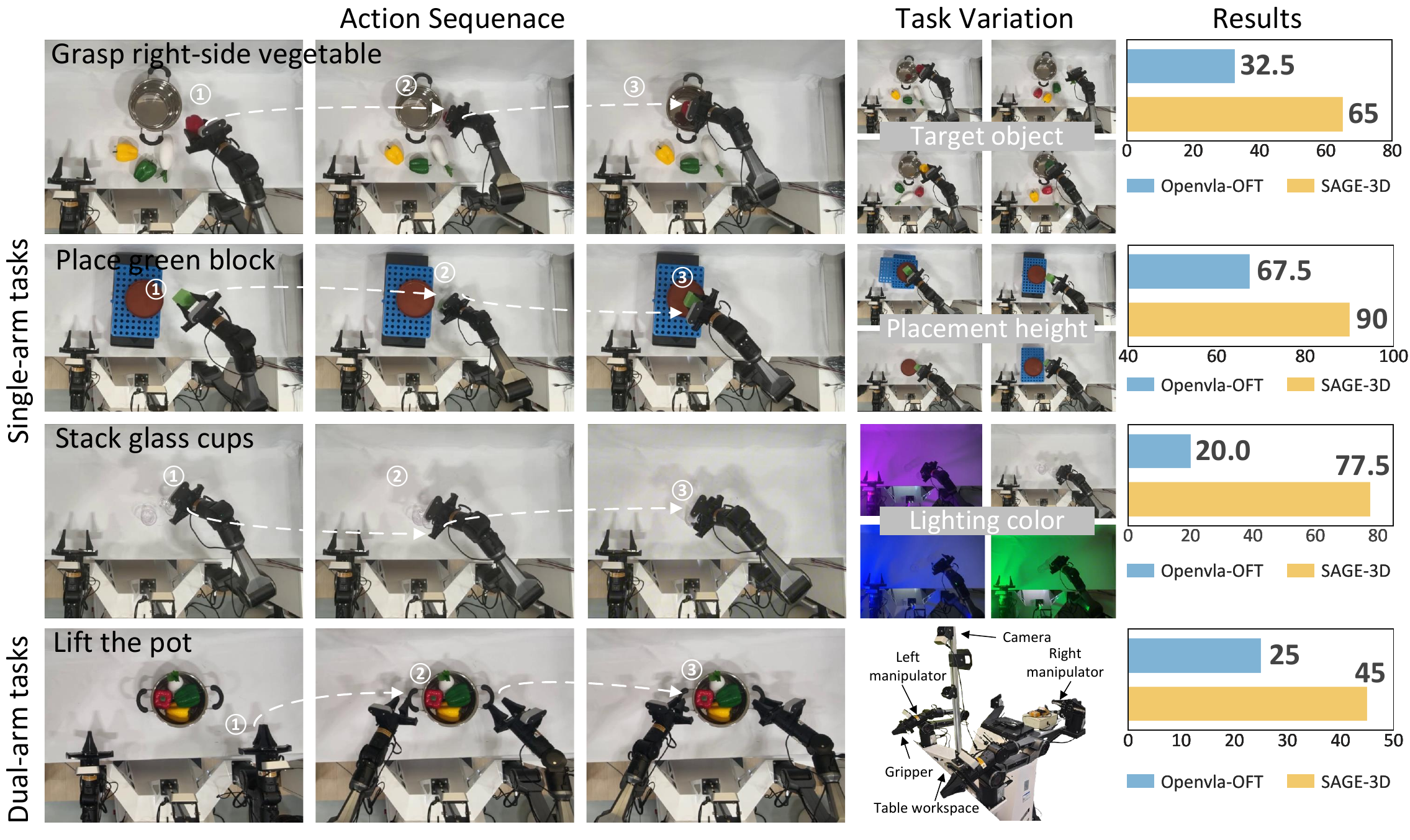}
  \caption{\textbf{Real-world robotic manipulation experiments.} \methodname\ performs single-arm and dual-arm manipulation tasks under real-world variations.}
  
  \label{fig:real}
  \vspace{-16pt}
\end{figure}

\textbf{Real-world evaluation.} Fig.~\ref{fig:real} reports the real-world results under four task variations. \methodname{} improves over OpenVLA-OFT on all tasks, from 32.5\% to 65\% on \textit{grasp right-side vegetable}, 67.5\% to 90\% on \textit{place green block}, 20\% to 77.5\% on \textit{stack glass cups}, and 25\% to 45\% on \textit{lift the pot}. These gains cover target-object changes, placement-height changes, lighting shifts, and dual-arm coordination, indicating that the injected semantic-aware 3D representation provides task-relevant spatial cues beyond 2D appearance features. The largest improvement appears under lighting variation, where SAGE-3D gains 57.5 percentage points, suggesting reduced reliance on appearance shortcuts. The improvement on the dual-arm task further shows that the learned 3D cues remain useful when action prediction requires object-level pose and balance reasoning.

%% file: sec/5_relat.tex
\vspace{-2pt}
\section{Related Work}
\vspace{-2pt}
\subsection{General Vision-Language-Action Policies}

 Vision-Language Models (VLMs) provide the semantic foundation for embodied intelligence through strong visual understanding and language reasoning capabilities~\cite{VLM,Prismatic,Molmo,Molmo2}. Vision-Language-Action Models (VLAs) build on these representations to map visual observations and language instructions to robot actions. Early works such as RT-1~\cite{RT1} and RT-2~\cite{RT2} demonstrated the potential of language-conditioned robot control, while later methods improved scalability, efficiency, and long-horizon manipulation through open-source VLM backbones, flow matching, and policy refinement~\cite{OpenVLA,openvla-oft,pi0,pi0.5,pi0.6,ABot-M0,MolmoAct,MolmoAct2,spiritv1.5,lin2025hif,song2025reconvla}. However, most existing VLAs rely primarily on 2D visual representations, which limits their ability to reason about the spatial structure of the 3D world in geometry-sensitive tasks.

\subsection{3D-aware Vision-Language-Action Policies}
Recent work augments VLAs with 3D information through two main paradigms. \emph{Explicit 3D fusion} incorporates depth maps, point clouds, or dedicated geometric encoders\cite{bhat20253d,4D-VLA,li2025bridgevla,cVLA,li2025pointvla,sun2025geovla,FP3}. \emph{Implicit 3D injection} learns spatial cues from RGB observations through multi-view projection, depth estimation, 3D representation learning, or teacher-guided feature alignment\cite{OG-VLA,qu2025spatialvla,DepthVLA,spatialforcing,FALCON,occvla,VidBot,LaMP}. Other methods enhance spatial reasoning with semantic 3D representations, spatial grounding, or structured reasoning modules\cite{InternVLA-M1,GraphCoT-VLA}. Nevertheless, existing approaches often treat geometry as an auxiliary signal without sufficiently aligning it with task-relevant semantics, or depend on scene-specific adaptation, limiting their robustness under distribution shifts. Our work addresses this gap through an explicit semantic-aware 3D representation for generalizable VLA policies.



%% file: sec/7_con.tex
\section{Conclusion and Limitations}

We present \textbf{\methodname{}}, an efficient and robust framework for improving 3D geometric reasoning in VLA models. By extracting semantic-aware 3D representations from SAFG and injecting them into the policy branch, \methodname{} provides task-relevant geometric priors while preserving the pretrained semantic knowledge of the underlying VLM. Experiments on \textsc{LIBERO}, \textsc{LIBERO-Pro}, and real-world tasks show consistent gains in both in-domain performance and cross-domain generalization.

\methodname{} introduces an additional 3D reasoning branch, incurring extra computation over standard 2D VLA models. We mitigate this overhead with lightweight feature generation and efficient token injection, and provide an SAFG pretrained on robot data as a reusable initialization for downstream manipulation tasks. However, monocular depth errors and single-view occlusions can impair 3D reasoning by obscuring geometry and object relations. Future work will explore distillation, feature compression, efficient 3D representations, and multi-view or sensor-based depth cues.



%% file: sec/appendix.tex
\newpage
\appendix
\section{Architecture Details of SAFG}\label{midi}

We provide the architectural details of SAFG. Unlike traditional depth estimators that only output geometric quantities, such as depth maps or point clouds, SAFG generates high-level semantic-aware 3D representations. SAFG comprises DepthPro~\cite{bochkovskiy2025depth}, CLIP ViT-L/14~\cite{clip}, SigLIP-SO400M~\cite{siglip}, and Mamba-130M~\cite{mamba}. It is pretrained on the million-scale SSR-CoT~\cite{liu2025ssr} corpus without robot actions or target-benchmark data. During policy training, only the two-layer GELU Alignment Encoder is optimized, while all other SAFG components remain frozen. The forward process of SAFG is defined as follows.

\textbf{Depth Prior Generation.}
Given an observation $\mathcal{O}_t \in \mathbb{R}^{H \times W \times 3}$ and a task instruction $\mathcal{T}$, we first use the pretrained monocular depth estimation model DepthPro to generate a high-resolution relative depth map $\mathcal{D}_t \in \mathbb{R}^{H \times W \times 1}$, which serves as an explicit geometric prior.

\textbf{Dual-Stream Feature Extraction and Alignment.}
We extract visual and depth features from the observation $\mathcal{O}_t$ and depth map $\mathcal{D}_t$, respectively. Specifically, we use a pretrained CLIP ViT-L/14 as the visual encoder $\mathcal{E}_{\mathrm{vis}}$ and SigLIP as the depth encoder $\phi_{\mathrm{depth}}$. To align heterogeneous visual and depth features into a unified semantic embedding space, we introduce an MLP projector with two fully connected layers and GELU activation. The feature extraction process is formulated as:
\begin{equation}
    Z_V = \mathcal{E}_{\mathrm{vis}}(\mathcal{O}_t), \quad
    Z_D = \phi_{\mathrm{depth}}(\mathcal{D}_t).
\end{equation}

\textbf{Efficient Language Model Spatial Reasoning.}
To jointly reason over visual semantics, depth geometry, and the task instruction, we introduce a efficient language model Mamba $\mathcal{M}$. It encodes $Z_V$, $Z_D$, and $\mathcal{T}$ to generate latent states $H_R$ that capture spatial relations and affordance-relevant cues:
\begin{equation}
    H_R = \mathrm{LastHidden}(\mathcal{M}(Z_V, Z_D, \mathcal{T})).
\end{equation}

\textbf{3D Reasoning Token Projection.}
Finally, we employ a latent projection module $\phi_{\mathrm{proj}}$ to map $H_R$ into the target embedding space used by the downstream VLA model. The final 3D reasoning tokens are computed as:
\begin{equation}
    Z_{\mathrm{RT}} = \phi_{\mathrm{proj}}(H_R) \in \mathbb{R}^{L \times D_{\mathrm{RT}}}.
\end{equation}

The resulting $Z_{\mathrm{RT}}$ is used by the downstream VLA policy, providing 3D spatial cues for precise physical manipulation while preserving the pretrained semantic knowledge of the original model.

\section{Supplementary Experiments}

\subsection{Setup Details of Simulation Benchmarks}

\noindent\textbf{LIBERO Training Setup:} 
Our base model is OpenVLA-OFT (7B Prismatic)~\cite{openvla-oft}, pretrained on the OXE dataset~\cite{open-X-Embodiment} with DINOv2~\cite{dinov2}, SigLIP~\cite{siglip}, and Llama-2~\cite{llama2} as its backbone components. We freeze the pretrained weights and optimize rank-32 LoRA adapters on all VLM linear layers, including the visual encoder, while training the proprioception projector and action head from scratch. We train SAGE-3D on a single NVIDIA H800 GPU using AdamW with a learning rate of $5e^{-4}$, cosine annealing scheduler with warm-up, and batch size 8. Detailed hyperparameter settings are provided in Table~\ref{tab:comparison}. We observed that training on LIBERO causes forgetting and conflicts for some tasks, so we set the max iteration value to be a multiple of the number of iterations corresponding to one epoch to ensure optimal performance. For \textsc{LIBERO-Long}, we train for 12 epochs, while for \textsc{LIBERO-Spatial}, \textsc{LIBERO-Object}, and \textsc{LIBERO-Goal}, training for 4 epochs is sufficient. The number of steps required to traverse the training set of each suite is summarized in Table~\ref{tab:training_steps}.

\begin{table}[!htbp]
\centering
\vspace{-8pt}
\caption{\textbf{Training hyperparameter settings of SAGE-3D on LIBERO.} We use the same settings across all LIBERO suites unless otherwise specified.}
\vspace{4pt}
\label{tab:comparison}
\setlength{\tabcolsep}{6pt}
\begin{tabular*}{0.95\linewidth}{@{\extracolsep{\fill}}l|cccc}
\toprule[0.5pt]
Setting & Batch size & Max training step & Learning rate & Warmup step \\
\midrule[0.5pt]
Value & 8 & 12 epochs & $5\times10^{-4}$ & 500 \\
\bottomrule[0.5pt]
\end{tabular*}
\vspace{-5pt}
\end{table}

\begin{table}[!htbp]
\vspace{-8pt}
\centering
\caption{\textbf{Per-epoch training steps for LIBERO suites.} Computed as dataset size divided by batch size (8) and validated during training.}
\vspace{4pt}
\label{tab:training_steps}
\setlength{\tabcolsep}{6pt}

\begin{tabular*}{0.95\linewidth}{@{\extracolsep{\fill}}l|cccc}
\toprule[0.5pt]
LIBERO Suite & \textsc{Spatial} & \textsc{Object} & \textsc{Goal} & \textsc{Long} \\
\midrule[0.5pt]
Steps Per Epoch & 6,622 & 8,373 & 6,506 & 12,684 \\
\bottomrule[0.5pt]
\end{tabular*}

\end{table}

\begin{figure}[H]
  \centering
  \includegraphics[width=1.0\linewidth]{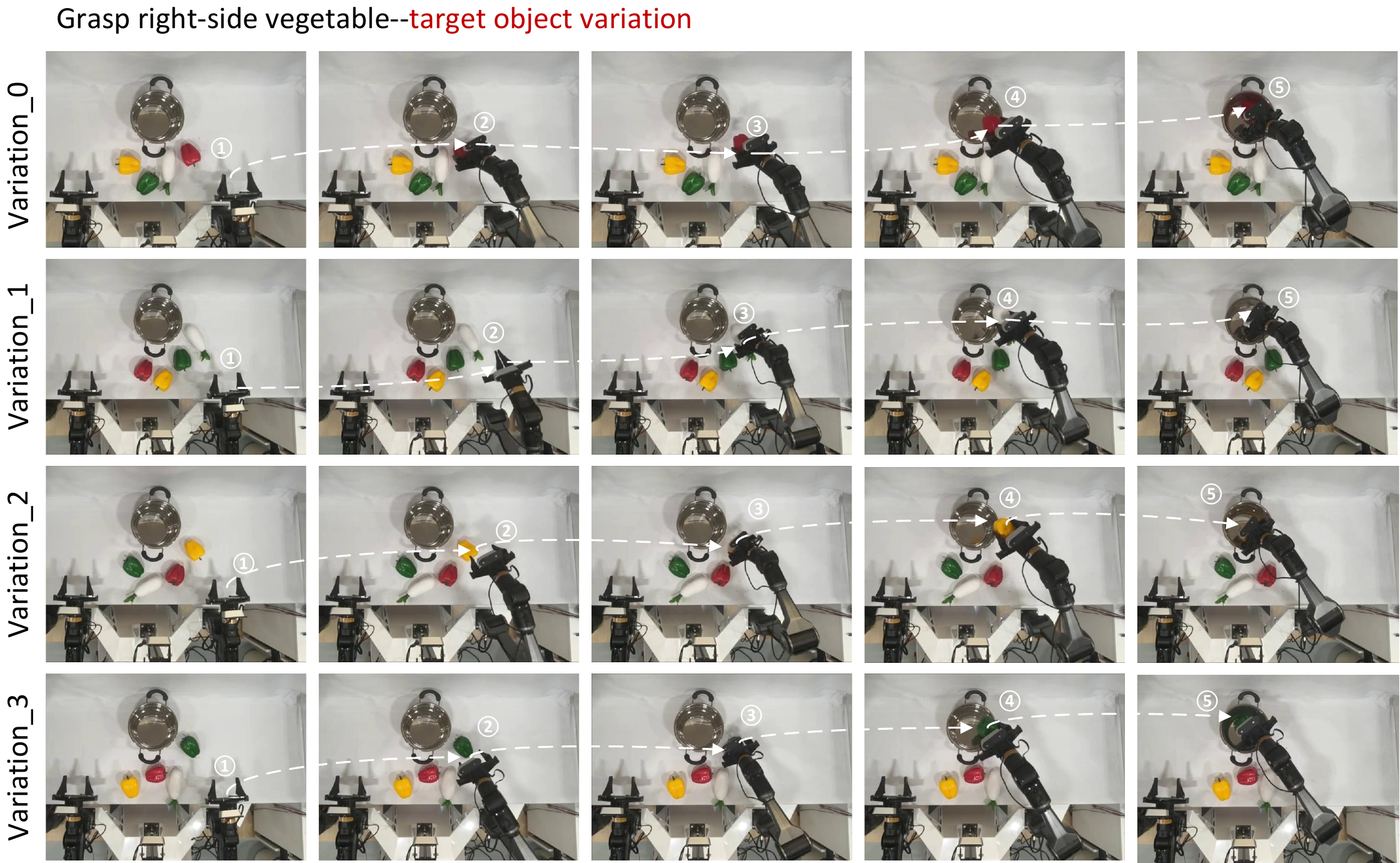} 
  \label{fig:real-1}
\end{figure}




\subsection{Performance on \textsc{LIBERO} Subtasks}
We present the detailed performance of \methodname{} on the 40 \textsc{LIBERO} subtasks in Table~\ref{tab:libero_details}. The results show that \methodname{} achieves consistently high success rates across all four \textsc{LIBERO} suites. In particular, it obtains perfect performance on all 10 \textsc{Spatial} subtasks and near-perfect performance on the \textsc{Object} and \textsc{Goal} suites, with average SRs of 99.8\% and 99.2\%, respectively. On the more challenging \textsc{Long} suite, which requires longer-horizon execution and is more susceptible to error accumulation, \methodname{} still achieves a strong average SR of 97.0\%. These fine-grained results indicate that the overall improvement is not driven by a few easy subtasks but is consistent across diverse manipulation scenarios.
\begin{table}[!htbp]
	\centering
	\setlength{\tabcolsep}{1mm}
	\caption{\textbf{Detailed success rates (\%) of \methodname{} on the 40 subtasks across four \textsc{LIBERO} suites.}}
    \label{tab:libero_details}
    \vspace{4pt}
	\begin{tabular}{l|cccccccccc|c}
		\toprule[1pt]
		\textbf{LIBERO} & \textbf{1} & \textbf{2} & \textbf{3} & \textbf{4} & \textbf{5} & \textbf{6} & \textbf{7} & \textbf{8} & \textbf{9} & \textbf{10} & \textbf{Avg.}  \\
		\midrule[1pt]
		Spatial & 100.0 & 100.0 & 100.0 & 100.0 & 100.0 & 100.0 & 100.0 & 100.0 & 100.0 & 100.0 & 100.0 \\
		Object  & 100.0 & 100.0 & 98.0 & 100.0 & 100.0 & 100.0 & 100.0 & 100.0 & 100.0 & 100.0 & 99.8 \\
		Goal    & 100.0 & 100.0 & 94.0 & 100.0 & 98.0 & 100.0 & 100.0 & 100.0 & 100.0 & 100.0 & 99.2 \\
		Long    & 100.0 & 100.0 & 100.0 & 94.0 & 100.0 & 100.0 & 92.0 & 100.0 & 90.0 & 94.0 & 97.0 \\
		\bottomrule[1pt]
	\end{tabular}%
    \vspace{-12pt}
\end{table}

\subsection{Details of Real-world Task Variations}

\begin{figure}[H]
  \centering
  \includegraphics[width=1.0\linewidth]{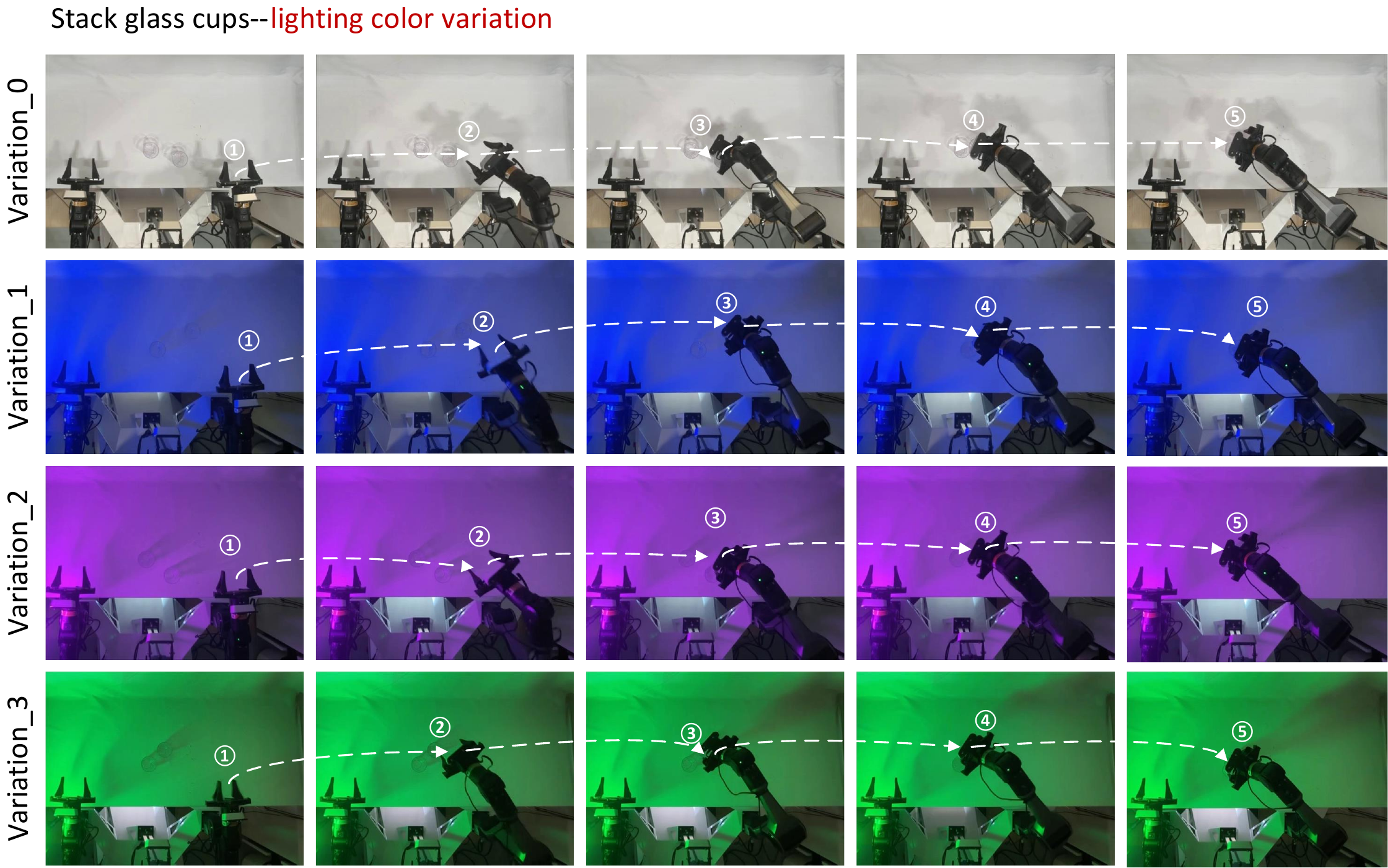}  
  \caption{\textbf{Action sequences of single-arm task variations.} We visualize the robot execution processes for three representative single-arm tasks under different real-world variations, including object position, height, and lighting conditions. }
  \label{fig:real-3}
\end{figure}

We provide additional details of the single-arm real-world experiments in Fig.~\ref{fig:real-3}, including task variations and their corresponding action sequences. This appendix focuses on three representative single-arm tasks, covering variations under different real-world settings, such as object position, height, and lighting conditions. For each task variation, we report the action sequence executed by the robot to complete the task, providing a more detailed illustration of the manipulation procedure.

\begin{figure}[H]

  \centering
  \includegraphics[width=1.0\linewidth]{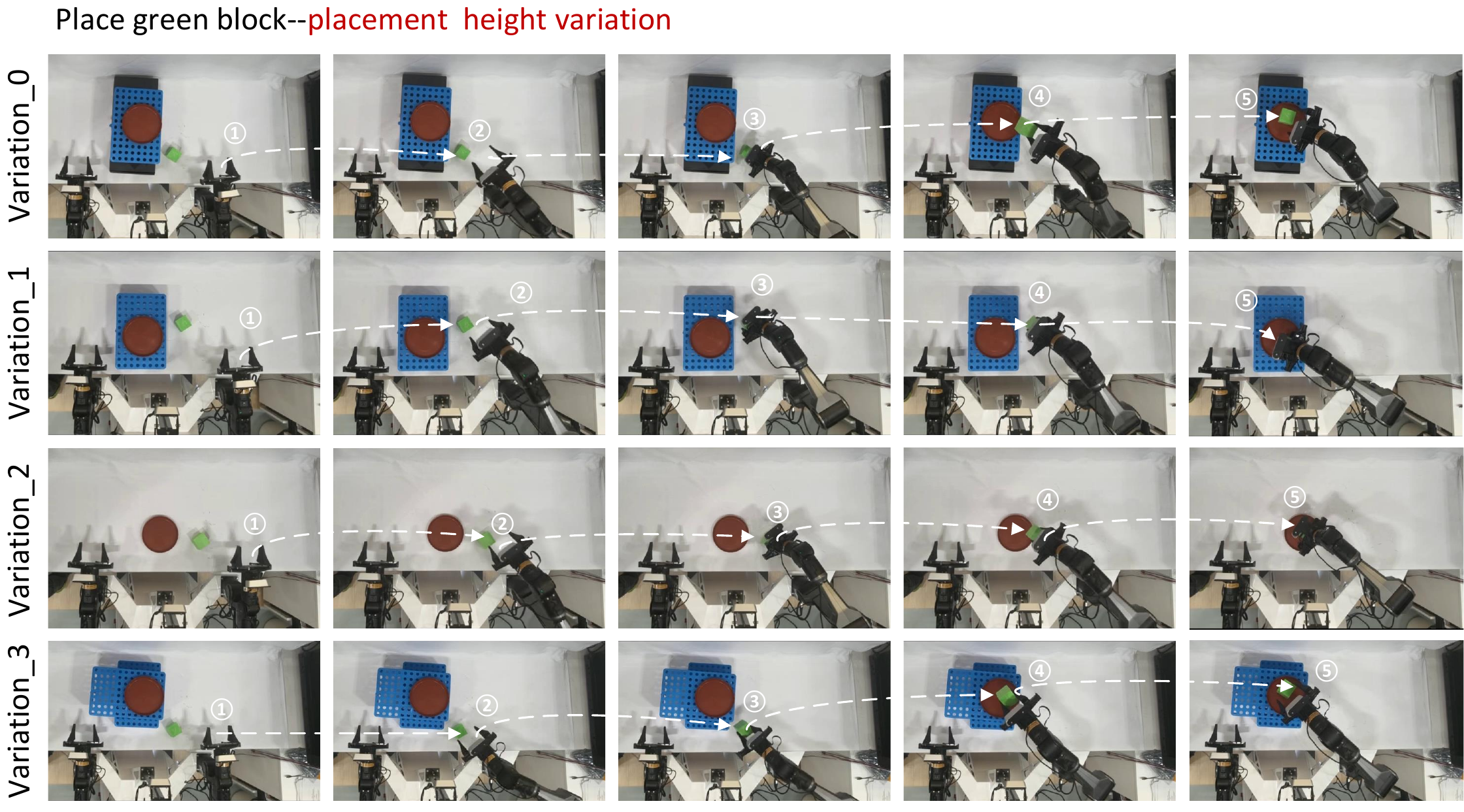} 
  \label{fig:real-2}
  \vspace{-20pt}
\end{figure}

\section{Execution Examples}
\begin{figure*}[h]
  \centering
  \includegraphics[width=1.0\linewidth]{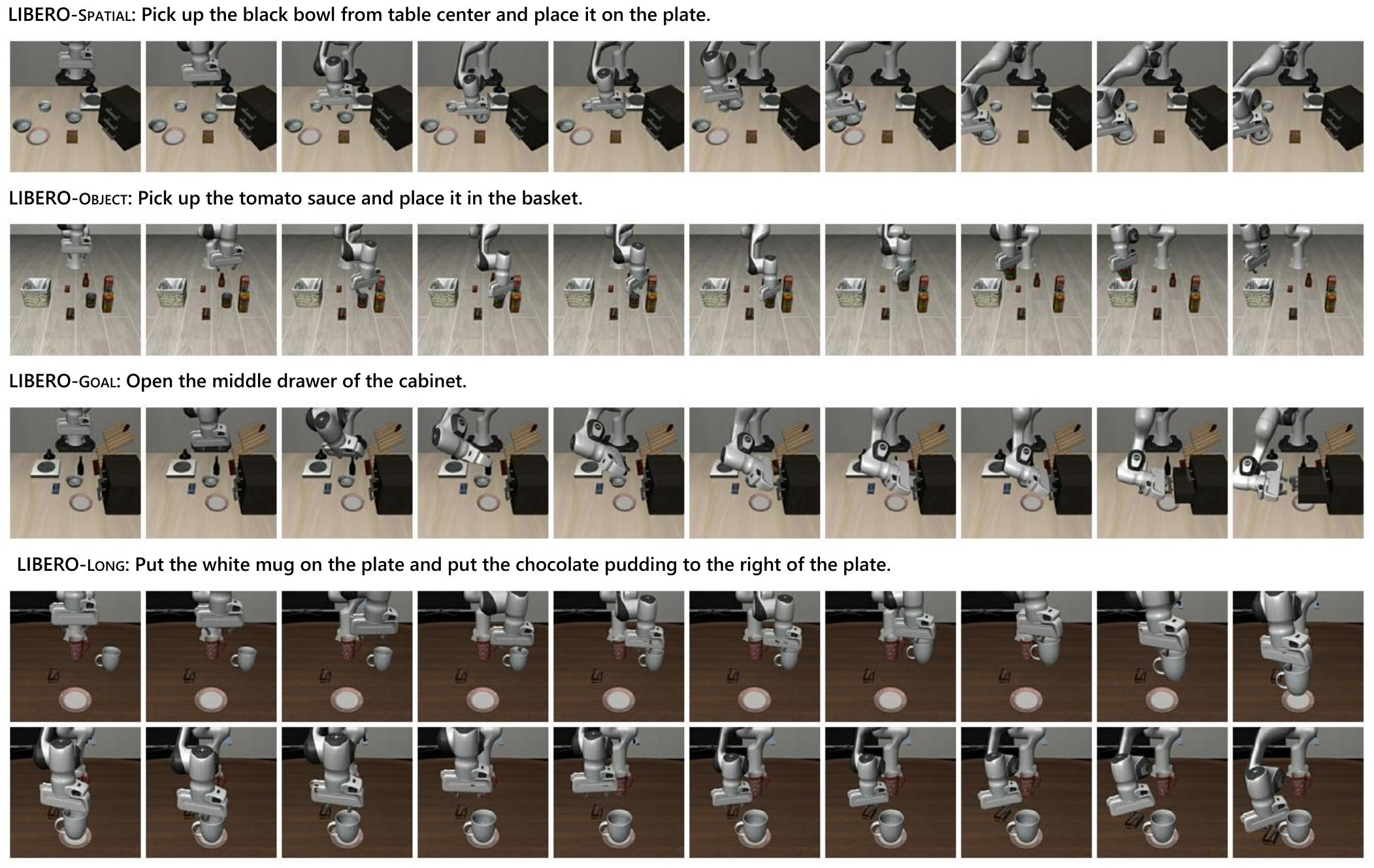}  
  \caption{\textbf{Representative execution trajectories on the LIBERO benchmark.} SAGE-3D successfully completes diverse manipulation tasks across four suites: \textsc{LIBERO-Spatial}, \textsc{LIBERO-Object}, \textsc{LIBERO-Goal}, and \textsc{LIBERO-Long}, demonstrating precise instruction following and temporally coherent action generation.}
  \label{fig:examples}
\end{figure*}

To evaluate the execution capabilities and robustness of our model across diverse scenarios, we visualize representative rollout sequences from the LIBERO benchmark in Fig.~\ref{fig:examples}. Specifically, we showcase execution examples across four distinct task categories: \textsc{LIBERO-Spatial}, which emphasizes spatial reasoning; \textsc{LIBERO-Object}, focusing on identifying specific target entities; \textsc{LIBERO-Goal}, requiring precise placement relative to environmental fixtures; and \textsc{LIBERO-Long}, which tests the model's performance in extended manipulation sequences. These examples demonstrate the model's ability to translate high-level linguistic instructions into precise, temporally coherent manipulation actions across diverse robotic tasks.

